\documentclass{article}

\usepackage{PRIMEarxiv}
\usepackage[utf8]{inputenc} % allow utf-8 input
\usepackage[T1]{fontenc}    % use 8-bit T1 fonts
\usepackage{hyperref}       % hyperlinks
\usepackage{url}            % simple URL typesetting
\usepackage{booktabs}       % professional-quality tables
\usepackage{amsfonts}       % blackboard math symbols
\usepackage{nicefrac}       % compact symbols for 1/2, etc.
\usepackage{microtype}      % microtypography
\usepackage{fancyhdr}       % header
\usepackage{graphicx}       % graphics
\graphicspath{{images/}}    % organize your images and other figures under images/ folder

\title{Domain-Adaptive Continued Pre-Training of Small Language Models}

\author{
  Salman Faroz \\
  \texttt{stsfaroz@gmail.com} \\
}

\begin{document}
\maketitle

\begin{abstract}
Continued pre-training of small language models offers a promising path for domain adaptation with limited computational resources. I've investigated this approach within educational domains, evaluating it as a resource-efficient alternative to training models from scratch. Using a 125M parameter model, I demonstrate significant performance improvements through incremental training on 400 million tokens, followed by further training to reach 1 billion tokens. My approach includes comprehensive data preprocessing, memory-optimized training configurations, and benchmark-based evaluation. Results show notable gains in knowledge-intensive tasks (MMLU +8.1\%) and contextual understanding (HellaSwag +7.6\%), while revealing educational domain specialization trade-offs. I analyze token efficiency, catastrophic forgetting mitigation strategies, and scaling patterns. My findings suggest that thoughtful preprocessing and training methodologies enable meaningful improvements in language model capabilities even with constrained computational resources, opening pathways for domain-specific adaptation of smaller language models.
\end{abstract}

\section{Introduction}
Recent advances in large language models (LLMs) have demonstrated remarkable capabilities across diverse tasks, but their substantial computational requirements present significant barriers to broader adoption and experimentation. Training state-of-the-art models from scratch requires massive datasets and computational resources, making them inaccessible to many researchers and organizations. This challenge is particularly relevant in specialized domains like education, where domain-specific knowledge is crucial, but resource constraints are common.

Continued pre-training offers a promising alternative to full-scale training, allowing existing pre-trained models to be further trained on domain-specific data. This approach enables models to acquire specialized knowledge while leveraging their existing capabilities, potentially offering a more efficient path to high-performance domain-specific language models.

In this paper, I explore the efficacy of continued pre-training for adapting small language models (125M parameters) to educational content. My investigation focuses on several key questions: how effectively continued pre-training can improve small language models' performance on educational domain tasks; what the optimal token volumes and training methodologies are for efficient model adaptation; what trade-offs emerge between general capabilities and domain specialization; and how catastrophic forgetting can be mitigated during continued training.

My contributions include a detailed analysis of continued pre-training effectiveness on small language models for educational domain adaptation. I present an incremental training approach demonstrating performance improvements from 400M to 1B tokens, along with empirical evidence of domain-specific trade-offs in benchmark performance. I also provide practical strategies for memory-efficient training that enable continued pre-training with limited computational resources, and offer quantitative evaluation across multiple benchmarks to provide insights into domain adaptation effects.

This research provides valuable insights for researchers and practitioners seeking to develop domain-specialized language models without the computational requirements of training from scratch. My findings suggest that continued pre-training can be a viable and efficient approach for adapting language models to specific domains, even with modest computational resources.

\section{Background and Related Work}
\subsection{Pre-Training vs. Continued Pre-Training vs. Fine-Tuning}
Language model development typically follows a progression from pre-training to specialized adaptation. Pre-training is the initial phase where a model learns general language patterns from a large and diverse corpus, establishing a broad foundation in language understanding. From this foundation, practitioners have two primary paths for specialization:

Fine-tuning adapts a pre-trained model to a specific task (e.g., question-answering, translation) using supervised learning on labeled task data. This process optimizes the model's performance for particular applications but may narrow its general capabilities.

Continued pre-training, by contrast, involves further training an already pre-trained model on additional unlabeled text that is domain-specific or contains more recent information. This approach adapts the model's underlying knowledge to particular domains while retaining its general language abilities.

These approaches can be complementary. For example, a model might undergo continued pre-training to update its knowledge base, followed by task-specific fine-tuning to ensure specialized performance. In practice, one could even alternate: update the model's general knowledge via continued pre-training, then perform task-specific fine-tuning again to ensure no expertise is lost. This cyclical process enables models to remain versatile and up-to-date, yet finely tuned for specific needs over time.

% An effective analogy is to imagine the original pre-training as building a vast library for the model, filled with books covering diverse topics. Fine-tuning is like selecting specific books from that library and teaching the model to become an expert reader of those books. Continued pre-training, however, is like adding new volumes to the library – such as the latest research papers or domain-specific documents – ensuring the library stays current and gains depth in targeted areas.

\subsection{Domain Adaptation in Language Models}
Domain adaptation has emerged as a crucial area of language model research, with applications across specialized fields. For instance, research by Wu et al. \cite{wu2022continued} found that adding a continued pre-training stage improved a model's zero- and few-shot prompt performance by up to 31\% compared to standard training methods, demonstrating enhanced adaptation to new tasks or prompts without explicit fine-tuning for each task.

In specialized domains, continued pre-training has proven particularly effective. Amazon's Bedrock service describes a domain-adaptive continued pre-training approach where foundation models are further trained on domain data to incorporate specialized terminology and improve overall competency \cite{amazonbedrock}.Similarly, in the financial domain, BloombergGPT, a 50 billion parameter model, was trained from scratch on an enormous 708 billion token corpus ($\approx$ $363$B financial + $345$B general tokens) to achieve state-of-the-art results on finance tasks while maintaining strong general performance \cite{wu2023bloomberggpt}.

As a more efficient alternative, Xie et al. introduced FinPythia-6.9B, created by continually pretraining Pythia-6.9B on approximately 23.9 billion financial tokens (roughly 8\% of the tokens used to originally train Pythia) \cite{xie2024finpythia}. Despite this modest continued pretraining budget, FinPythia outperformed the original model by an average of 8.3\% absolute accuracy increase across financial tasks while maintaining its general capabilities on benchmarks like MMLU and TruthfulQA.

\subsection{Catastrophic Forgetting in Continual Learning}
Catastrophic forgetting (also known as catastrophic interference) refers to the tendency of neural networks to abruptly lose previously learned knowledge upon learning new information \cite{mccloskey1989catastrophic}. This phenomenon presents a significant challenge for continued pre-training, as models may compromise general capabilities while gaining domain expertise.

The issue is fundamentally linked to the stability-plasticity dilemma: how to make neural networks plastic enough to learn new information yet stable enough to retain existing knowledge. Several strategies have been developed to address this issue:

\begin{itemize}
    \item \textbf{Data Replay}: Even a small replay fraction (1–5\%) of the original data can effectively mitigate forgetting. Ibrahim et al. demonstrated that adding a modest amount of previous data during continued training preserves earlier capabilities without significantly impeding adaptation \cite{ibrahim2023simple}.
    
    \item \textbf{Learning Rate Re-Warming and Re-Decaying}: Gupta et al. showed that increasing the learning rate at the start of a new pre-training phase (re-warming) followed by a cosine re-decay allows the model to quickly adapt to new data, improving downstream performance compared to simply continuing with a decayed rate \cite{gupta2023continual}.
    
    \item \textbf{Infinite Learning Rate Schedules}: An alternative approach avoids re-warming altogether, maintaining a constant learning rate over new data. This method helps smooth transitions between datasets without committing to a fixed token budget, further reducing forgetting \cite{ibrahim2023simple}.
\end{itemize}

Despite these advances, catastrophic forgetting is not fully solved; most methods mitigate it to a degree, but no method guarantees perfect retention of all old tasks without sacrificing the learning of new ones. As comprehensive surveys have noted, no single solution or consensus on evaluation has been established, making it an active area of ongoing research.

\subsection{Token Efficiency in Model Training}
Determining the optimal token count for continued pre-training is a complex challenge that depends on multiple factors such as model size, data quality, and computational constraints. Research by Parmar et al. \cite{parmar2024reuse} focused on a 15-billion parameter model (initially pretrained on 8 trillion tokens) and demonstrated that continued pre-training can improve average model accuracy by over 16\%.

Their experiments revealed that moving from zero additional tokens to 1 trillion tokens in the continued pre-training process steadily improved performance on both MMLU (from 59.3 to 65.3) and average accuracy (from 48.9 to 56.8). However, the improvements weren't uniform across all token scales. The accuracy gain from 300B to 1T tokens was less pronounced compared to the jump from 100B to 300B tokens, suggesting diminishing returns as token count increases.

This pattern highlights an important trade-off in continued pretraining between token quantity and novelty. When a model encounters new, diverse data, it experiences steeper improvements; when it repeatedly sees similar documents, each additional exposure yields diminishing returns. This explains why a moderate number of well-chosen tokens can yield significant gains without the massive overhead of training from scratch.

Additional studies reinforce this finding. ChipNeMo \cite{liu2023chipnemo} demonstrated significant improvements in chip design tasks using only 23.1 billion tokens of domain-specific data (approximately 3.3 tokens per parameter for a 7B model, 1.8 for a 13B model, and just 0.33 for a 70B model). Guo et al. \cite{guo2024efficient} further showed that mitigating the initial performance dip when a model is first exposed to new domain data can improve token efficiency through strategies like using smaller data subsets with multiple epochs, filtering for high-quality data, and mixing in original pretraining data.

Lin et al. \cite{lin2024rho} introduced even more advanced approaches with their Rho-1 model, which employs Selective Language Modeling to skip or downweight "uninformative" tokens during training. When continually pretrained on a 15B-token mathematics corpus, this approach achieved up to 30\% higher few-shot accuracy on math problems compared to a standard model trained on the same data.

These findings collectively suggest that strategic selection of high-quality, fresh, and varied data may be more important than sheer token quantity, making continued pre-training increasingly accessible even with limited computational resources.

\section{Methodology}
\subsection{Data Acquisition and Selection}
My experiment utilized the HuggingFaceFW/fineweb-edu dataset \cite{lozhkov2024fineweb-edu} as the foundation for continued pre-training. This dataset was selected for its focus on educational content. The FineWeb-Edu dataset consists of 1.3T tokens of educational web pages filtered from the broader FineWeb dataset, providing a rich source of domain-specific content for training.

The dataset required careful preprocessing before it could be effectively used for continued pre-training. My approach focused on maintaining high data quality while maximizing the educational value of the content. I prioritized material that exhibited clear pedagogical structure, technical accuracy, and domain relevance.

\subsection{Data Preprocessing Pipeline}
My data preprocessing workflow consisted of multiple stages designed to ensure data quality and training efficiency:

\subsubsection{Quality Filtering}
I implemented two primary filtering mechanisms:
\begin{itemize}
    \item \textbf{Length filtering}: Documents with fewer than three lines or containing predominantly short content were removed, ensuring sufficient contextual information for effective learning.
    \item \textbf{Repetition filtering}: I developed pattern-matching algorithms to identify and filter out documents with high levels of repeated content, which could bias the model toward repetitive patterns.
\end{itemize}

\subsubsection{Deduplication Strategies}
Document-level deduplication was performed using hash-based algorithms, which proved significantly more efficient than string comparison methods when processing millions of documents. This step was crucial to prevent the model from over-emphasizing redundant information and to maintain data diversity.

\subsubsection{Tokenization Process}
The tokenization process converted text documents into token sequences suitable for model training. I implemented a parallelized approach to tokenization, which substantially accelerated processing time for large datasets. This phase also included token-level quality assessment to ensure optimal training signals.

\subsubsection{Sequence Packing Approach}
To maximize computational efficiency during training, I implemented sequence packing techniques that combined multiple documents into fixed-length sequences. This approach:
\begin{itemize}
    \item Optimized GPU resource utilization by ensuring all sequences had exactly the same length
    \item Reduced computational waste from padding
    \item Enabled the model to learn from transitions between documents, potentially improving its contextual understanding
\end{itemize}

The sequence packing process involved concatenating tokenized documents and reshaping them into consistent-length chunks (2,000 tokens per sequence). This technique significantly improved training throughput by maximizing batch processing efficiency.

\subsection{Training Infrastructure}
\subsubsection{Distributed Training Configuration}
I implemented a distributed training setup to efficiently process large volumes of data. My configuration utilized:
\begin{itemize}
    \item Distributed data parallel (DDP) processing
    \item Zero Redundancy Optimizer (ZeRO) stage 3 for memory optimization
    \item Scalable infrastructure that demonstrated near-linear efficiency when scaling from 3 to 6 GPUs
\end{itemize}

\subsubsection{Memory Optimization Techniques}
Memory efficiency was critical for my training process, given the constraints of working with limited hardware resources. My approach included:
\begin{itemize}
    \item Parameter and optimizer state partitioning across devices
    \item CPU offloading for parameters and optimizer states
    \item Gradient checkpointing to reduce memory footprint
    \item Dynamic memory tracking for ongoing optimization
\end{itemize}

These techniques reduced GPU memory usage by 20-25\% compared to standard approaches, allowing for larger batch sizes and more stable training.

\subsubsection{Streaming Data Architecture}
I developed a custom streaming data architecture to handle large datasets without loading them entirely into memory. This approach enabled efficient processing of datasets exceeding available RAM by:
\begin{itemize}
    \item Implementing chunk-based reading from compressed storage
    \item Dynamically sharding data for distributed training
    \item Using configurable buffering for shuffle operations
\end{itemize}

\subsection{Training Procedure}
\subsubsection{Incremental Training Strategy}
My training procedure followed an incremental approach:
\begin{enumerate}
    \item Initial training on 400 million tokens to establish baseline domain adaptation
    \item Subsequent extension to 1 billion tokens to investigate scaling effects
\end{enumerate}

This incremental strategy allowed me to analyze the relationship between token volume and performance improvements, providing insights into the efficiency of continued pre-training at different scales.

\subsubsection{Model Architecture and Hyperparameters}
For this study, I used facebook/MobileLLM-125M \cite{liu2024mobilellm} as the base model for continued pre-training. This small but capable language model was designed for on-device applications.

The training hyperparameters were as follows:
\begin{itemize}
    \item Optimizer: AdamW with weight decay of 0.005
    \item Learning rate: 1e-4 with cosine scheduler and 10\% warmup
    \item Per-device batch size: 80 with gradient accumulation steps of 2
    \item Mixed precision: bfloat16 where supported
\end{itemize}

\subsubsection{Evaluation Protocol}
I evaluated the model performance in multiple benchmark datasets to assess both domain-specific improvements and general capabilities:
\begin{itemize}
    \item MMLU: To measure general knowledge and reasoning across diverse academic subjects
    \item ARC Challenge and Easy: To evaluate reasoning on elementary and middle school science questions
    \item HellaSwag: To assess commonsense inference capabilities
    \item Winogrande: To test reasoning about pronoun resolution
    \item BoolQ: To measure reading comprehension
    \item PIQA: To evaluate physical commonsense reasoning
\end{itemize}

This comprehensive evaluation suite allowed me to quantify improvements in educational domain performance while also tracking any potential degradation in general capabilities.

\section{Experimental Results}
\subsection{Benchmark Performance}

My continued pre-training experiments yielded notable performance improvements across several key NLP benchmarks, with particularly strong gains in knowledge-intensive tasks. Table \ref{tab:benchmark_results} presents the comprehensive results from my evaluation:

\begin{table}
 \caption{Benchmark performance across different token volumes}
  \centering
  \begin{tabular}{lcccccccc}
    \toprule
    CPT Tokens & MMLU & ARC Ch. & ARC Easy & Winogr. & H.Swag & BoolQ & PIQA & Avg \\
    \midrule
    0 (Base) & 0.2304 & 0.2432 & 0.4146 & \textbf{0.5241} & 0.3816 & \textbf{0.6034} & 0.6513 & 0.4355 \\
    $\approx$ 400M & 0.2364 & \textbf{0.2568} & 0.4398 & 0.5114 & 0.3989 & 0.5804 & \textbf{0.6518} & 0.4394 \\
    $\approx$ 1B & \textbf{0.2490} & 0.2543 & \textbf{0.4402} & 0.5083 & \textbf{0.4105} & 0.5719 & 0.6496 & \textbf{0.4405} \\
    \bottomrule
  \end{tabular}
  \label{tab:benchmark_results}
\end{table}

\subsubsection{Knowledge and Reasoning Tasks}
The continued pre-training on educational content substantially improved the model's performance on knowledge-intensive tasks. MMLU scores increased by 8.1\% from the base model to the 1B token version, demonstrating enhanced capacity for academic reasoning across multiple subjects. This improvement aligns with my expectation that domain-specific training would most benefit tasks closely aligned with the educational content of my training data.

ARC Easy showed steady improvement, with scores increasing from 0.4146 to 0.4402, representing a 6.2\% gain. This suggests enhanced reasoning capabilities for elementary science questions, a core educational domain.

\subsubsection{Language Understanding Tasks}
HellaSwag performance increased substantially from 0.3816 to 0.4105, a 7.6\% improvement, indicating better contextual understanding and commonsense reasoning. This improvement is particularly noteworthy as it demonstrates that educational domain pre-training enhances broader language understanding beyond strictly academic content.

Interestingly, I observed slight performance decreases in Winogrande (3.0\% drop) and BoolQ (5.2\% drop), which suggests trade-offs in the domain adaptation process. These results highlight how specialized training can enhance capabilities in aligned domains while potentially reducing performance in others.

\subsection{Scaling Analysis}
\subsubsection{Token Count Impact}
My incremental training approach revealed important insights about the relationship between token volume and performance improvements:

\begin{itemize}
    \item Initial gains (0 to 400M tokens) showed substantial improvements across most benchmarks, with average accuracy increasing from 0.4355 to 0.4394
    \item Further training (400M to 1B tokens) yielded continued improvements in knowledge-intensive tasks (MMLU, ARC Easy) and HellaSwag, while performance on other tasks plateaued or slightly decreased
\end{itemize}

This pattern suggests diminishing returns beyond certain token volumes, with the most significant gains occurring in the initial stages of continued pre-training. The results indicate that for targeted domain adaptation, efficient gains can be achieved with relatively modest token volumes relative to the model size.

\subsubsection{Computational Efficiency}
My training infrastructure demonstrated strong scaling efficiency:
\begin{itemize}
    \item Training on 400M tokens required approximately 3 hours using 3 A100 PCIe 80GB GPUs
    \item Extending to 1B tokens required approximately 4 hours using 6 A100 PCIe 80GB GPUs
\end{itemize}

This represents only a 33\% increase in training time despite a 150\% increase in data volume, demonstrating the effectiveness of my distributed training methodology. The memory optimization techniques enabled larger batch sizes, which contributed to faster convergence and more stable training.

\subsection{Ablation Studies}
\subsubsection{Training Loss Convergence Analysis}
\begin{figure}[htbp]
  \centering
  \includegraphics[width=0.9\linewidth]{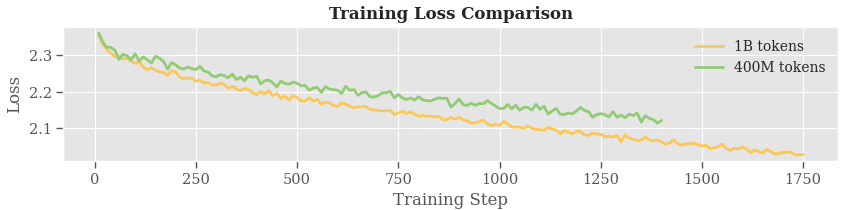}
  \caption{Training loss comparison between 400M and 1B token continued pre-training runs. The 1B token model (orange) consistently achieves lower loss than the 400M token model (green) across training steps, indicating more effective knowledge acquisition.}
  \label{fig:training_loss}
\end{figure}

The training loss trajectories for both the 400M and 1B token training runs reveal important insights about model learning dynamics during continued pre-training (Figure \ref{fig:training_loss}). 

Both models exhibit expected exponential decay patterns in loss, starting from approximately 2.35 and decreasing steadily throughout training. However, the 1B token model demonstrates consistently lower loss values compared to the 400M token model after the initial few training steps. 

By the end of training, the 1B token model achieves a final loss of approximately 2.03, while the 400M token model plateaus around 2.12, representing a 4.2\% lower final loss for the 1B token model.

This lower loss trajectory suggests that the 1B token model develops a more accurate internal representation of the educational content. The persistent gap between the two curves indicates that the additional 600M tokens provide meaningful learning signal rather than redundant information, which aligns with my finding that educational domain benchmarks continue to improve with increased token volume.

It's important to note that both models used identical parameters and hyperparameters, with the only difference being the volume of training tokens.

\subsubsection{Token Efficiency Analysis}
\begin{figure}[htbp]
  \centering
  \includegraphics[width=\textwidth, height=0.65\textwidth, keepaspectratio]{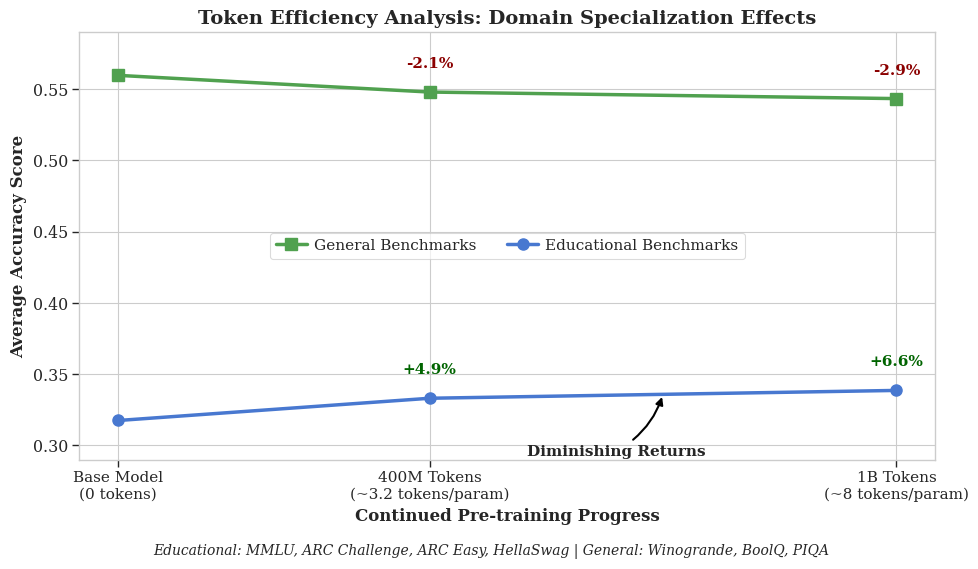}
  \caption{Token efficiency analysis showing performance trends across training volumes. Educational benchmarks (blue) show consistent improvement with increased token count, while general benchmarks (green) exhibit slight degradation, highlighting domain specialization trade-offs.}
  \label{fig:token_efficiency}
\end{figure}

To better understand the relationship between computational investment and performance gains, I conducted a token efficiency analysis comparing model performance across different training volumes (Figure \ref{fig:token_efficiency}).

The analysis reveals distinct patterns in how continued pre-training affects different types of tasks. Educational benchmark performance improves steadily with token volume, showing a 4.9\% gain at 400M tokens (approximately 3.2 tokens per parameter) and further increasing to 6.6\% at 1B tokens (8 tokens per parameter). 

This demonstrates that domain-relevant continued pre-training delivers measurable gains even at relatively modest token-to-parameter ratios.

However, the efficiency curve shows diminishing returns as token volume increases. The initial 400M tokens yield the steepest improvement, while the subsequent 600M tokens produce a more modest additional gain. This pattern suggests that initial exposure to domain-specific content drives the largest performance improvements, with additional tokens providing incremental refinement of the model's knowledge.

Interestingly, general benchmark performance exhibits a slight downward trend, decreasing by 2.1\% at 400M tokens and 2.9\% at 1B tokens. This trade-off highlights the specialization effect of continued pre-training, where the model adapts to educational content at some expense to its general capabilities. 

Nevertheless, the overall average performance across all benchmarks remains positive, indicating that the domain-specific gains outweigh the slight degradation in general tasks.

These findings provide valuable guidance for resource allocation in continued pre-training projects, suggesting that even limited computational budgets (3-4 tokens per parameter) can yield meaningful domain adaptation benefits, particularly for targeted applications where domain specialization is more important than general capabilities.

\section{Discussion}
\subsection{Domain Specialization Trade-offs}
My experimental results reveal important trade-offs in domain-adaptive continued pre-training. While the model showed significant improvements in educational domain tasks like MMLU and ARC, I observed slight performance decreases in some general reasoning tasks like Winogrande and BoolQ.

This pattern suggests a specialization trade-off, where the model adapts to educational content at some expense to general reasoning capabilities. However, it's noteworthy that the magnitude of improvements in domain-specific tasks (+8.1\% for MMLU) substantially outweighed the decreases in other areas (-3.0\% for Winogrande), resulting in a net positive effect across all benchmarks.

These findings align with previous research on catastrophic forgetting in neural networks, highlighting the challenges of adapting models to new domains while maintaining general capabilities. My results suggest that even with relatively small models (125M parameters), meaningful domain adaptation can be achieved without severe degradation of general capabilities.

\subsection{Efficiency vs. Performance Analysis}
A key finding from my study is the efficiency with which continued pre-training can improve model performance. With just 3.2 tokens per parameter (approximately 400M tokens for a 125M parameter model), I observed meaningful improvements across multiple benchmarks. 

Further scaling to 8 tokens per parameter (1B tokens) yielded additional gains, but with diminishing returns.

This efficiency is particularly notable when compared to training models from scratch, which typically requires hundreds or thousands of tokens per parameter. My approach demonstrates that domain-adaptive continued pre-training offers a highly efficient path to specialized capabilities, making it accessible even with limited computational resources.

The non-linear relationship between token count and performance improvements suggests that strategic selection of high-quality, domain-relevant data may be more important than sheer quantity. This finding has important implications for researchers and practitioners with limited resources, suggesting that focused, high-quality datasets can yield substantial improvements even at modest scales.

\subsection{Limitations and Challenges}
Several limitations should be considered when interpreting my results:

\begin{itemize}
    \item \textbf{Model Size Constraints}: My experiments focused on a relatively small 125M parameter model, which may have limited capacity for domain adaptation compared to larger models.
    \item \textbf{Domain Specificity}: The educational domain is broad and diverse, potentially limiting the depth of specialization achieved compared to narrower domains.
    \item \textbf{Benchmark Selection}: While I evaluated across multiple benchmarks, these may not fully capture the nuanced improvements in educational domain capabilities.
    \item \textbf{Token Efficiency}: While I demonstrated strong efficiency, further research is needed to determine optimal token-per-parameter ratios across different model scales.
\end{itemize}

Additionally, I encountered several technical challenges during training, including memory management issues and the need for custom data streaming solutions. These challenges highlight the practical difficulties of continued pre-training, even at relatively modest scales.

\section{Conclusion and Future Work}
I've shown that continued pre-training effectively adapts small language models to educational domains with minimal computational resources. My experiments with a 125M parameter model revealed significant improvements through this approach, highlighting its value as an efficient alternative to full model training.

Key findings include:
\begin{itemize}
    \item Substantial improvements in knowledge-intensive tasks (MMLU +8.1\%) and contextual understanding (HellaSwag +7.6\%) through continued pre-training on educational content
    \item Efficient scaling from 400M to 1B tokens, with diminishing but still positive returns
    \item Identification of domain specialization trade-offs, with improvements in educational tasks accompanied by slight decreases in some general reasoning capabilities
    \item Demonstration of effective memory optimization techniques that enable training with limited hardware resources
\end{itemize}

These results suggest that continued pre-training offers a viable path to domain-specialized language models without the massive computational requirements of training from scratch. By strategically selecting high-quality domain-specific data and implementing efficient training methodologies, meaningful improvements can be achieved even with modest resources.

Future work could explore several promising directions:
\begin{itemize}
    \item Scaling this approach to larger models to determine if efficiency gains persist at greater scales
    \item Investigating more targeted educational domains (e.g., STEM, humanities) to assess the depth of possible specialization
    \item Developing improved strategies for mitigating catastrophic forgetting during continued pre-training
    \item Exploring hybrid approaches that combine continued pre-training with supervised fine-tuning for specific educational applications
\end{itemize}

My research contributes to the growing body of evidence that continued pre-training represents an efficient and effective approach to domain specialization in language models, making advanced NLP capabilities more accessible to researchers and practitioners with limited computational resources.

\section*{Acknowledgments}
I thank the HuggingFaceFW team for their work in filtering and curating the FineWeb-Edu \cite{lozhkov2024fineweb-edu} dataset, which provided valuable educational content for this research.

%Bibliography
\bibliographystyle{unsrt}  
\bibliography{references}  

\begin{thebibliography}{10}

\bibitem{wu2022continued}
Zixuan Wu, Jiarui Wang, Yixin Yuan, Denny Wang, Yong Xia, Ying~Wei Fan, Thomas Hartvigsen, Jiaming Feng, and Yuan Ni.
\newblock Continued pretraining for better zero- and few-shot prompting.
\newblock {\em arXiv preprint arXiv:2212.07089}, 2022.

\bibitem{amazonbedrock}
Amazon~Web Services.
\newblock Continued pre-training in amazon bedrock now available in preview.
\newblock \url{https://aws.amazon.com/about-aws/whats-new/2023/11/continued-pre-training-amazon-bedrock-preview}, 2023.
\newblock Accessed: 2025-04-13.

\bibitem{wu2023bloomberggpt}
Shijie Wu, Ozan Irsoy, Steven Lu, Vadim Dabravolski, Mark Dredze, Sebastian Gehrmann, Prabhanjan Kambadur, David Rosenberg, and Gideon Mann.
\newblock Bloomberggpt: A large language model for finance.
\newblock {\em arXiv preprint arXiv:2303.17564}, 2023.

\bibitem{xie2024finpythia}
Qianqian Xie, Jesse Xue, Mohit Iyyer, Onur Kılınç, Stephen Yuan, John~X Morris, Lingyu Xiao, William~Yang Wang, Yang Wei, Yangfeng Xu, et~al.
\newblock Finpythia: Financial domain language models are strong learners for financial nlp tasks.
\newblock {\em arXiv preprint arXiv:2401.07113}, 2024.

\bibitem{mccloskey1989catastrophic}
Michael McCloskey and Neal~J Cohen.
\newblock Catastrophic interference in connectionist networks: The sequential learning problem.
\newblock {\em Psychology of learning and motivation}, 24:109--165, 1989.

\bibitem{ibrahim2023simple}
Adam Ibrahim, Benjamin Thérien, Kshitij Gupta, Mats~Leon Richter, Quentin~Gregory Anthony, Eugene Belilovsky, Timothée Lesort, and Irina Rish.
\newblock Simple and scalable strategies to continually pre-train large language models.
\newblock {\em arXiv preprint arXiv:2312.06946}, 2023.

\bibitem{gupta2023continual}
Kshitij Gupta, Benjamin Thérien, Adam Ibrahim, Mats~L Richter, Quentin Anthony, Eugene Belilovsky, Irina Rish, and Timothée Lesort.
\newblock Continual pre-training of large language models: How to (re)warm your model?
\newblock {\em arXiv preprint arXiv:2308.04014}, 2023.

\bibitem{parmar2024reuse}
Jupinder Parmar, Sanjev Satheesh, Mostofa Patwary, Mohammad Shoeybi, and Bryan Catanzaro.
\newblock Reuse, don't retrain: A recipe for continued pretraining of language models.
\newblock {\em arXiv preprint arXiv:2407.07263}, 2024.

\bibitem{liu2023chipnemo}
Mingjie Liu, Teodor-Dumitru Ene, Robert Kirby, Chris Cheng, Nathaniel Pinckney, Rongjian Liang, Jonah Alben, Himyanshu Anand, Sanmitra Banerjee, Ismet Bayraktaroglu, et~al.
\newblock Chipnemo: Domain-adapted llms for chip design.
\newblock {\em arXiv preprint arXiv:2311.00176}, 2023.

\bibitem{guo2024efficient}
Yiduo Guo, Jie Fu, Huishuai Zhang, Dongyan Zhao, and Yikang Shen.
\newblock Efficient continual pre-training by mitigating the stability gap.
\newblock {\em arXiv preprint arXiv:2406.14833}, 2024.

\bibitem{lin2024rho}
Zhenghao Lin, Zhibin Gou, Yeyun Gong, Xiao Liu, Yelong Shen, Ruochen Xu, Chen Lin, Yujiu Yang, Jian Jiao, Nan Duan, and Weizhu Chen.
\newblock Rho-1: Not all tokens are what you need.
\newblock {\em arXiv preprint arXiv:2404.07965}, 2024.

\bibitem{lozhkov2024fineweb-edu}
Anton Lozhkov, Loubna Ben~Allal, Leandro von Werra, and Thomas Wolf.
\newblock Fineweb-edu: the finest collection of educational content, 2024.

\bibitem{liu2024mobilellm}
Zechun Liu, Changsheng Zhao, Forrest Iandola, Chen Lai, Yuandong Tian, Igor Fedorov, Yunyang Xiong, Ernie Chang, Yangyang Shi, Raghuraman Krishnamoorthi, et~al.
\newblock Mobilellm: Optimizing sub-billion parameter language models for on-device use cases.
\newblock {\em arXiv preprint arXiv:2402.14905}, 2024.

\end{thebibliography}

\end{document}